\pdfoutput=1
\documentclass[11pt]{article}

\usepackage[final]{acl}

\usepackage{times}
\usepackage{latexsym}
\usepackage{tablefootnote}
\usepackage{hyperref}

\usepackage[T1]{fontenc}
\usepackage[utf8]{inputenc}

\usepackage{microtype}

\usepackage{inconsolata}

\usepackage{graphicx}
\usepackage{subcaption}
\usepackage{graphicx}
\usepackage{soul}
\usepackage{xurl}
\usepackage{amsmath}

\title{Inducing lexicons of in-group language\\ with socio-temporal context}

\author{Christine de Kock
  \\
  University of Melbourne
  \\
  Computing and Information Systems
  \\
  \texttt{christine.dekock@unimelb.edu.au}
  }

\begin{document}
\maketitle
\begin{abstract}
In-group language is an important signifier of group dynamics. This paper proposes a novel method for inducing lexicons of in-group language, which incorporates its socio-temporal context. Existing methods for lexicon induction do not capture the evolving nature of in-group language, nor the social structure of the community. Using dynamic word and user embeddings trained on conversations from online anti-women communities, our approach outperforms prior methods for lexicon induction. We develop a test set for the task of lexicon induction and a new lexicon of manosphere language, validated by human experts, which quantifies the relevance of each term to a specific sub-community at a given point in time. Finally, we present novel insights on in-group language which illustrate the utility of this approach.
\end{abstract}

\section{Introduction}\label{sec:intro}
Social groups have a tendency to develop distinctive vernaculars, also known as \textit{in-group language}. According to \citet{drake1980social}, in-group language plays two important social roles within groups: first, it can obscure the discussions of the group from out-group onlookers or moderators, and second, it signals cohesion and solidarity with a group by illustrating an awareness and acceptance of their norms. Since this type of informal language evolves rapidly \citep{stewart2018making}, the use of recent lexical innovations of the group is a strong signal of group belonging \citep{danescu2013no,zhang2017community}.

Due to its strong relation to implicit social dynamics, lexicons of in-group language are frequently used to study communities (e.g. \citealp{baele2023diachronic, rowe2016mining}). Studies that use hand-compiled lexicons are dependent on the skill of the lexicon builders \citep{havaldar2024building}. In the case of in-group language, there are subtle signals that are challenging to identify as an out-group member \citep{mendelsohn2023dogwhistles}, especially given the scale of modern text corpora. Since manual lexicon building is expensive, there is a tendency not to update lexicons frequently. For example, \citet{rowe2016mining} include a 7-year-old lexicon in their study on online pro-ISIS language. These factors impact the quality of lexicons, with resulting negative effect on the quality of downstream studies.

Prior work in NLP has proposed computational methods for inducing lexicons of in-group language, often framing the task as identifying relevant words using large text corpora from the group in question. For example, \citet{lucy2021characterizing} use statistical properties of word occurrences and contextualised word embeddings. Similarly, \citet{farrell2020use} and \citet{yoder2023identity} use approaches based on topic models and static word embeddings. However, existing methods for computationally identifying in-group language tend to focus only on linguistic information, despite the highly dynamic and social nature of groups and their language.

In this work, we propose and evaluate methods for {\textbf{L}}exicon {\textbf{I}}nduction with {\textbf{S}}ocio-{\textbf{T}}emporal {\textbf{N}}uance (\textbf{LISTN}) using dynamic word- and user embeddings. This framework provides a natural way to integrate statistical properties of the text with the social structure of the community, by modelling users and words in the same space. Our experiments indicate that LISTN-based methods outperform a range of baselines based on prior work, achieving an average precision score of 0.77 on a newly crafted test set for lexicon induction\footnote{Github: \url{github.com/unimelb-nlp/listn}}.

Our study centers on \textit{the manosphere}, a growing movement of anti-women online communities which has been described as a violent extremist ideology \citep{baele2023diachronic} and has been associated several real-world acts of terror against women \citep{latimore2023incels}. These communities are exceptionally linguistically innovative \citep{bogetic2023race}, making them a particularly appropriate application area for this work, in addition to being a pressing current concern. The manosphere is a highly fractured collective, with well-developed and continuously evolving sub-group structure \citep{ribeiro2021evolution}, which also highlights the need for incorporating social and temporal dimensions for lexicon induction. 

To evaluate our lexicon induction methodology, we use existing lexicons of manosphere language and human expert validation to construct a novel test set. We further release a lexicon of 455 new manosphere terms with scores representing their relevance to specific manosphere subcommunities. Using this lexicon, we find that word embeddings of in-group terms tend to be more static (i.e.\ showing less variation over time) compared to words of similar frequencies in the general vocabulary. 
Finally, we look at linguistic specialisation within subgroups of the manosphere, finding that Incels are the most distinctive in their vernacular. These analyses illustrate the utility of our approach for studying in-group language as a phenomenon, and for drawing out latent group dynamics as signalled through language.

\section{Related work}
A number of prior works have investigated in-group language, its evolution over time, and its relation to social dynamics. \citet{chancellor2016thyghgapp} investigate linguistic variation and the processes by which existing words are modified to create neologisms in the context of pro-anorexia groups, finding that lexical innovations are associated with increased engagement and more extreme content. \citet{danescu2013no} explored the role of linguistic evolution in groups, using snapshot n-gram language models to measure the deviation of a user from the community. They illustrate that the non-adoption of lexical innovations is predictive of user churn. \citet{zhang2017community} characterises how distinctive and temporally dynamic online communities are in their language, finding that communities with more dynamic and distinctive language are likely to have higher retention of members. \citet{stewart2018making} and \citet{del-tredici-fernandez-2018-road} investigate factors that influence the dissemination of new lexical innovations, finding that linguistic and social factors both play a role in this process. These studies illustrate the intimate relationship between social dynamics and language, and the dynamic nature of in-group language and groups. Our work uses a holistic approach for incorporating these different effects to induce lexicons of in-group language. 

Language cues have been used to investigate extremist movements; for instance, by \citet{baele2023diachronic} and \citet{de2024investigating} in the context of the manosphere, and by \citet{rowe2016mining} to study ISIS radicalisation. However, these works rely on small, manually constructed lexicons. Through this work, we hope to enable future research on this important topic by producing lexicons with a strong empirical justification, grounded in the socio-temporal context of the community being studied.

Our methods leverage dynamic, jointly-learned word and user embeddings (described in Section \ref{sec:method}). The representation learning methodology was introduced in our earlier study, \citet{de2024jointly}, to forecast what communities a user will interact with, to characterise the level of violent language within sub-groups of the manosphere, and to measure changes in the relevance of temporal concepts (e.g. \textit{\#MeToo}) to the community. Here, we reimplement and optimise the methodology (described in Section \ref{sec:implementatoin} but use the same training data. We then propose and compare 6 different methods for using these embeddings to induce lexicons.

\section{Representation learning}\label{sec:method}
The Cerberus architecture \citep{de2024jointly} performs a shared temporal matrix factorisation over \textbf{content and adjacency} matrices, which are user-content and user-user representations. A \textbf{content-only} model variation decomposes user-content matrices over time to yield dynamic word embeddings.

\paragraph{Content-only} At each timestep $t$, a user-content matrix $C_t$ is constructed using positive pointwise mutual information (PPMI), which quantifies how strongly the presence of user $i$ and word $j$ are associated beyond what would be expected if they were independent (based on a background corpus):
\begin{align}\label{eq:ppmi}
    \text{PMI}(i,j) &= \text{log}\frac{P(j|i)}{P(j)}, \\
    \text{PPMI}(i,j) &= \text{max}\big(\text{PMI}(i,j),0\big).
\end{align}

$P(j)$ calculated is based on word frequencies in a background corpus of 40 million Reddit posts by \citet{dziri2019augmenting}. Matrix factorisation is used to learn user representations $U_t$ and word representations $W_t$ such that $C_t\approx U_t\cdot W_t^T$, for all $t$. 

\paragraph{Content and adjacency} To represent interpersonal dynamics, a user-user adjacency matrix $A_t$ is constructed to capture how frequently two users interact in the same thread at time $t$, with L1 normalisation. This matrix is jointly decomposed with $C_t$ to yield $U_t$, $V_t$ and $W_t$ such that $C_t\approx U_t\cdot W_t^T$ and $A_T\approx U_t\cdot V_t^T$, where $V_t$ is the context user embedding matrix. 

To ensure alignment of embeddings across timesteps, the learning objective includes a regularisation term that penalises large inter-timestep changes. The joint factorisation was found to outperform the content-only approach in \citet{de2024jointly}. 

\subsection{Data}\label{sec:data}
The models are trained on over 4 million utterances from 50 manosphere subreddits over 9 months, using the Reddit portion of the corpus of \citet{ribeiro2021evolution}. The subreddits in the dataset are mapped to five different manosphere sub-communities, referred to as \textbf{categories} in this paper: Incels, Men's Rights Activists (\textbf{MRA}), Men Going Their Own Way \textbf{MGTOW}, Pick-up Artists (\textbf{PuA}), and The Red Pill (\textbf{TRP}). The dataset further includes data from two anti-manosphere communities (referred to as \textbf{Criticism}) and three \textbf{Mental Health} related communities, which we use as a control group in this study. Embeddings are produced in monthly windows, representing April-December 2018. 

The data is filtered to include only users who are active for 3 or more timesteps and words that are used by more than 20 users, yielding 33,880 users and 44,679 words. The NLTK word tokeniser is used for tokenisation. Words are lowercased and non-alphabetic characters are removed.

\subsection{Implementation}\label{sec:implementatoin}
The models used in \citet{de2024jointly} were implemented in TensorFlow to match the evaluation setting of antecedent systems, and were not optimised for scaling to large datasets. We re-implement\footnote{Github: \url{github.com/unimelb-nlp/listn}} these models in PyTorch, using the generalised matrix factorisation formulation \citep{he2017neural} which (1)~avoids reading the full sparse matrices into memory, (2)~enables batch-wise updates for faster convergence, and (3)~allows for increased flexibility to deploy other models for representation learning.

In this formulation, a training step is performed over a batch of users for a given timestep, rather than updating only once all the data has been seen. During the forward pass, the model computes interaction vectors for the users in the batch against the full set of context users and words for that timestep. The model parameters are updated after each tuple of (\textit{user-user}, \textit{user-word}) batches. Hyperparameters are tuned based on the validation loss and are shown in Table \ref{tab:hypers} in Appendix \ref{app:hypers}.

\section{LISTN}\label{sec:listn}
Using the models described in Section~\ref{sec:method}, we evaluate a number of approaches for inducing lexicons. We refer to these methods as \textbf{LISTN} ({\textbf{L}}exicon {\textbf{I}}nduction with {\textbf{S}}ocio-{\textbf{T}}emporal {\textbf{N}}uance) and distinguish between \textbf{LISTN-CA} (jointly training on \textbf{C}ontent and \textbf{A}djacency matrices) and \textbf{LISTN-C} (using only the \textbf{C}ontent matrix).

Recall that the user-content matrix $C$ is constructed using PMI to compare how often a user uses a word compared to a large background corpus. We evaluate several methods for lexicon induction that are based on the low-rank reconstruction $\hat{C}_t$. The rationale is that we aim to find words that are used more often than expected (as indicated by PMI), using embeddings to provide a more nuanced view of a word or user compared to word frequencies. 

Given a user embedding $u_{i,t}$ and a word embedding $w_{j,t}$, we calculate the \textit{relevance} of word $j$ to user $i$ at time $t$ as ${r(i,j,t) = u_{i,t}\cdot w_{j,t}^T}$ (as also done in \citealp{de2024jointly}). This can be generalised to groups of users by finding the group centroid as the mean of the individual user embeddings, denoted $\overline{u}_t$. \citet{farrell2020use} find that the usage of in-group terms varies between the different manosphere categories. Therefore, we explore several methods that allow for sub-group specialisation. Each of the below methods produce a (scalar) score $r$ representing the relevance of a word $j$ within the manosphere ecosystem at time $t$.

\paragraph{Community centroid:} We calculate word relevance scores as ${r_{comm}(j,t) = \overline{u}_t\cdot w_{j,t}^T}$, where $\overline{u}_t$ represents the centroid of all users in the training set at time $t$.
\paragraph{Category:} We calculate a centroid $\overline{u}_{k,t}$ for each of the manosphere categories (as described in Section \ref{sec:method}), defined as the mean embedding of all users who contributed to category $k$ at timestep $t$. For every word, we then find its maximum relevance score across all categories:  ${r_{cat}(j,t) = \max_{k} (\overline{u}_{k,t}\cdot w_{j,t}^T)}$. We use maximum aggregation (as opposed to mean or median) because a word only needs to be highly relevant to one sub-group to be considered as valid in-group language.
\paragraph{Subreddit:} Similar to the category-based method, we find the maximum relevance score for a subreddit $p$ using its centroid: ${r_{sub}(j,t) = \max_{p} (\overline{u}_{p,t}\cdot w_{j,t}^T)}$.
\paragraph{Cluster:} We use K-means clustering of the user embeddings to find subcommunities. The motivation is that there may be groups that span different categories or subreddits, or sub-groups within a given category. For each cluster $n$ we find ${r_{clust}(j,t) = \max_{n} (\overline{u}_{n,t}\cdot w_{j,t}^T)}$. We do this for $K\in\{5,20,100\}$. Since the clustering depends on random initialisation, we report the mean over 5 runs.
\paragraph{Bootstrap:} Departing from the reconstruction-based approaches, we use a list of confirmed terms from manosphere lexicons (provided in Table~\ref{tab:lexicons}) and find their nearest neighbours based on cosine similarity of the $W$ embeddings. The score for each neighbouring word is given by its maximum similarity to a lexicon word.
\paragraph{Bias:} The factorisation models described in Section \ref{sec:method} include an inferred bias term for each user and word to capture population-level trends. In the case of the word bias, it captures the tendency for a word to have a high PMI across all users as compared to a background corpus. We use this directly as the word score.

\paragraph{Aggregation} These approaches produce a score for each word at every timestep. To aggregate, we use the maximum score of a word across all timesteps. This has the benefit of potentially capturing words that move into and out of popularity over the training window.

\section{Evaluation}\label{sec:evaluation} The goal of this work is to induce lexicons of in-group language. For the purpose of evaluation, we frame this as a \textbf{binary classification task} over \textbf{single-token lexical innovations}. Our rationale for this task definition and our evaluation setting is provided below. The test set constructed for this task is described in Section~\ref{sec:lexicons}.

A challenge in evaluating lexicons is that they are not a precise science and often rely on subjective expert judgment \citep{havaldar2024building}, as discussed in Section~\ref{sec:intro}. For example, the lexicons in Table~\ref{tab:group_terms} contain words such as \textit{hurt} and \textit{dumb}, which represent borderline cases in deciding whether they are sufficiently relevant for inclusion. A further challenge is that dogwhistles (words that take on a special meaning within an in-group that is different from its meaning to the general population) are difficult for researchers to identify \citep{lucy2021characterizing,mendelsohn2023dogwhistles}.

To address these challenges, we focus on \textbf{lexical innovations} in our evaluation. Lexical innovations are a form of in-group language which involve non-standard or extragrammatical words \citep{slotta2016slang}; for example, \textit{foid} is a common derogatory term for a woman in the Incel community. Lexical innovations provide a clearer decision boundary as they tend to be more easily attributed to a specific community. Importantly, this is a constraint on the evaluation and not a limitation of the method; all methods used in this work can produce relevance scores for standard and non-standard language alike.

A set of standard words is required in order to identify non-standard lexical innovations. For this purpose, we use the vocabulary of GloVe-400k \citep{pennington2014glove}, which is based on news and Wikipedia articles. Words that are not in this vocabulary are considered non-standard. We make an exception to include in-vocabulary words from existing manosphere lexicons (detailed in Section \ref{sec:lexicons}) in our evaluation, irrespective of whether they are lexical innovations. Since prior work has deemed these terms to be of particular importance to the manosphere, we consider them to be relevant and useful for evaluation.

Finally, we consider only \textbf{single-token words} to limit the vocabulary size. Future work may investigate N-gram models, since multi-word in-group expressions are also observed in the data. Within the manosphere, common multi-word expressions are often made into acronyms; for example, \textit{AWALT} is an acronym of ``all women are like that''.

\subsection{Baselines}\label{sec:baselines}
We use baselines based on two prior lexicon induction approaches:

\paragraph{Lexicon expansion} Similar to the approaches of \citet{yoder2023identity}, \citet{farrell2020use} and \citet{havaldar2024building}, we train a \texttt{word2vec} model \citep{mikolov2013distributed} on the same subset of posts used to train the \texttt{LISTN} models. We then find the nearest neighbours of known manosphere in-group language (based on the lexicons in Table \ref{tab:lexicons}) using cosine similarity over the $W$ embeddings. 

\paragraph{Statistical measures} Similar to \citet{lucy2021characterizing} and \citet{zhang2017community}, we evaluate variants of PMI, including PPMI (Eq. \ref{eq:ppmi}) and normalised PMI \textbf{(NPMI)}. Given a word $j$ and a particular setting (e.g.\ subreddit or community) $k$:
\begin{align}
    \text{NPMI}(j,k) &= \frac{\text{PMI}(j,k)}{-\text{log}P(j, k)}.
\end{align}

To determine $P(j)$ in the PMI calculation, we use the same background corpus used to construct the $C$ matrices. Probabilities are calculated based on word frequencies. Mirroring the \texttt{LISTN} evaluations, we account for sub-group specialisation by evaluating each PMI metric at the level of the full community, subreddit, and category, using the maximum score for a given word across the set of subgroups in each case. We also evaluate the NPMI and PPMI using monthly slices of data and maximum aggregation to mimic the temporally-aware nature of the \texttt{LISTN} methods. 

\subsection{Metrics} We report the average precision (AP) and area under the receiver-operator characteristic curve (AUROC) for each method. The AP provides a measure of the model's precision over all possible recall values; that is, how well the model avoids false positives. AUROC is a measure of its ability to distinguish between positive and negative cases. \citet{davis2006relationship} show that a model dominates in AUROC space if and only if it dominates in AP space, whereas the converse is not true; as such, we prioritise the AP. We use the randomised permutation test with Monte Carlo approximation ($R=9999$ and $\alpha=0.05$) for significance testing.

\section{Test set construction}\label{sec:lexicons}
We perform a preliminary scoring step for the purpose of constructing a test set for our main evaluation. Existing lexicons of manosphere language are used as a reference point. 

\paragraph{Existing lexicons}
The lexicons and the number of words in each, filtered to include only words that are in our corpus, are shown in Table~\ref{tab:lexicons}. For the \citet{farrell2019exploring} dataset, we follow \citet{ribeiro2021evolution} in excluding the lexicon categories that are not specific to the manosphere (e.g. racism). The Hatebase\footnote{\url{hatebase.org/}} and IncelWiki\footnote{\url{incels.wiki/w/Incel_Glossary}} lexicons were collected by us and will be released. Combined, this provides a set of 483 unique words.

\begin{table}[]
    \centering
    \begin{tabular}{|l|l|}
    \hline
        \textbf{Source} & \# \textbf{words} \\
        \hline
        IncelWiki & 189\\
        \citet{farrell2019exploring} & 194 \\
        \citet{baele2023diachronic} & 122 \\
        \citet{lucy2021characterizing} & 52 \\
        {Hatebase} & 29\\
        \hline
        \textbf{Total (deduplicated)} & \textbf{483} \\
        \hline
    \end{tabular}
    \caption{Existing lexicons used in the evaluation.}
    \label{tab:lexicons}
\end{table}

\paragraph{Preliminary scoring}
In this step, we evaluate each of the approaches in Sections~\ref{sec:listn}~and~\ref{sec:baselines} against the lexicons in Table~\ref{tab:lexicons}. Terms in the lexicons are assigned a positive label; all others are considered negatives. The average precision and area under the receiver-operator characteristic curve of each method are provided in Appendix~\ref{app:s1}. An issue with this approach, however, is that the lexicons suffer from low recall (a known issue; see \citealp{lucy2021characterizing}), meaning that true in-group terms identified by the methods are often wrongly evaluated as false positives. For this reason, we use this scoring only to identify words to include in our test set. 

\paragraph{Manual validation}
We retrieve the top-scoring 1000 terms produced by the best baseline and best \texttt{LISTN} approaches from the preliminary scoring step. These terms are manually evaluated by the author of this paper as well as a social psychology PhD student who is a specialist on the manosphere. Annotators are tasked with assigning a binary label to each word, based on whether or not it strongly signals membership of the manosphere. Annotators are also provided with three example usages of each word and the corresponding subreddit name. Following the expert annotator's advice, we also annotate ``cultural exports'' -- terms that originated within the manosphere but have spread to be used more widely. In our evaluation, we treat these terms as positives. We further include the terms from the existing lexicons in Table~\ref{tab:lexicons} as positive samples.

The resulting test set consists of 1803 words with binary labels, with an inter-annotator agreement score of 0.726 using Cohen's Kappa, indicating substantial agreement. In cases of disagreement, we use the label from the expert annotator. The test set is reasonably balanced, consisting of 944 positive and 859 negative samples. 

\section{Results}\label{sec:results}
\begin{table}[t]
    \centering
    \begin{tabular}{|l|l|l|}
        \hline & \textbf{AP} & \textbf{AUROC} \\
        \hline
        Random & 0.52& 0.5 \\
        \hline
        \multicolumn{3}{|c|}{\textbf{Baselines}}\\
        \hline
        \texttt{word2vec} bootstrap & 0.5563 & 0.5427\\
        PPMI & 0.6333&0.6709 \\
        PPMI-subreddit & 0.5910 	&0.5916 \\
        PPMI-category & 0.6517 &0.6470 \\
        PPMI-month &0.6165& 0.6071 \\
        NPMI & 0.6709 & \textbf{0.7034} \\
        NPMI-subredit & 0.6170 &	0.6187 \\
        NPMI-category & \textbf{0.6790} &0.6647 \\
        NPMI-month & 0.6449 & 0.6386\\

        \hline
        \multicolumn{3}{|c|}{\textbf{LISTN-CA}}\\
        \hline
        Community centroid  &0.6228&0.5131\\
        Category centroid &0.7231 &0.6713\\
        Subreddit centroid &0.5519 &0.5043\\
        Cluster-5 &\textbf{0.7620}&\textbf{0.7403}\\
        Cluster-20 & 0.7069&0.6954 \\
        Cluster-100&0.6950&0.6566\\
        Bootstrap &0.5349 &0.5206\\
        Bias &0.6190 &0.5560\\
        \hline
        \multicolumn{3}{|c|}{\textbf{LISTN-C}}\\
        \hline
        Community centroid & 0.6297&0.5255\\
        Category centroid&0.7272&0.6809\\
        Subreddit centroid&0.5891&0.5461\\
        Cluster-5 &\textbf{0.7679}&\textbf{0.7363}\\
        Cluster-20 & 0.7554&0.7289 \\
        Cluster-100&0.7040&0.6864\\
        Bootstrap & 0.5276&0.5077 \\
        Bias & 0.6016&0.5383 \\
        \hline
        \end{tabular}
        \caption{Comparison of different lexicon induction approaches.}
        \label{tab:results}
    \end{table}
Results are shown in Table~\ref{tab:results}. The best \texttt{LISTN-CA} and \texttt{LISTN-C} methods both outperform the best baseline method (statistically significant, $P<<0.05$). Using \texttt{LISTN-C} produces a higher AP score compared to \texttt{LISTN-CA} for 6 out of the 8 induction methods, including the top-performing method (not statistically significant; $P=0.723$). This is a surprising result, as it indicates that the induction does not benefit from the added author adjacency information (i.e.\ shared thread interaction) in \texttt{LISTN-CA}. A benefit of this outcome is that the \texttt{LISTN-C} model has fewer parameters and is more computationally efficient to train. The \texttt{LISTN-C} model does still contain social information, in that the source matrix is a user-content representation, meaning that users are represented as being similar if they use the same words. 

\paragraph{LISTN} For both \texttt{LISTN-CA} and \texttt{LISTN-C}, the 5-cluster approach performs the best ($AP=0.75$ and $0.77$ respectively). Using a larger number of clusters ($K=20$ and $K=100$) leads to worse performance. A possible explanation might be that smaller clusters lead to less meaningful centroids that overemphasise irrelevant words. On the other hand, averaging over the full community performs poorly, indicating that it is necessary to take into account \textit{some} level of sub-group linguistic specialisation. This is corroborated by the relatively low scores of the bias and bootstrap methods, which also do not account for community structure.

Notably, the 5-cluster method performs better than category-level induction, which computes centroids of a similar level of granularity. Subreddit-level induction (a more granular version of the category-level method, with 52 subgroups) performs the worst of the centroid-based approaches for both models and metrics. This is an unexpected result; one would expect this approach to perform on-par with the other centroid-based approaches. The cluster-based methods are intended to capture social structure at the intra-subreddit and inter-subreddit level, and assign each user to one cluster only. The subreddit representations average over all users who have contributed to a given subreddit per timestep, meaning that one user may form a part of many representations. This might include users who are incidental posters, which would add noise to the centroids. This effect would be more pronounced at the subreddit level than the category level, as there is less inter-category activity, which would explain why the category aggregation is better than subreddit aggregation but still worse than the clustering-based methods. Using a user weighting based on post frequency might resolve this issue. However, since the subreddit structure does not translate to other platforms, it is beneficial that the best method does not require it.

\paragraph{Baselines} All baseline approaches outperform the random baseline. The worst-performing baseline is the \texttt{word2vec}-based method. For both \texttt{LISTN} models, the lexicon expansion approaches (``bootstrap'') perform the worst as well. A possible reason for this might be that the seed lexicons are not varied enough, such that they are missing some area of the embedding space. The NPMI baseline performs the best on AUROC, whereas the category-level NPMI performs the best for the AP. In all cases, NPMI outperforms the PPMI, which corresponds with the results of \citet{lucy2021characterizing}. Interestingly, using month-level NPMI scoring does not improve over the time-aggregated NPMI, indicating that these superficial methods for accounting for time are not sufficient.

\section{Embedding analysis}
Beyond the improved performance over prior lexicon induction methods, an advantage of this approach is that it produces dynamic embeddings for words and users, in the same space. In this section, we use these embeddings to investigate (\textit{i}) changes in words over time and (\textit{ii}) sub-group linguistic specialisation in the manosphere.

\subsection{Change in word representations}
\begin{figure}
    \centering
    \includegraphics[width=\linewidth]{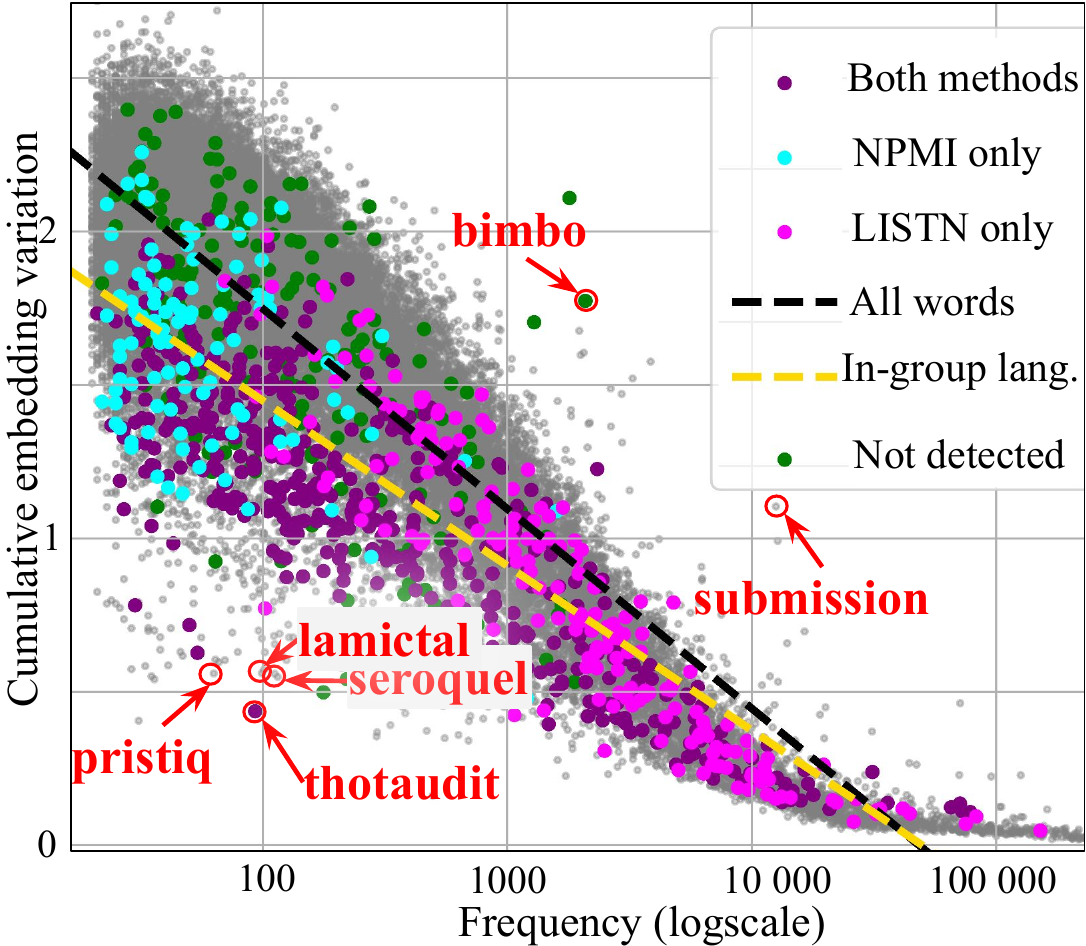}
    \caption{Word frequency versus embedding stability.}
    \label{fig:word-change}
\end{figure}

Recall that the training objective includes a regularisation term that penalises large changes between temporally successive word representations, meaning that changes in embeddings are only permitted if they reduce the reconstruction error sufficiently. In other words, word embeddings change as much as is needed, but no more. To determine how much a word changes between two successive timesteps, we calculate the cosine distance between successive $W_t$ representations. The sum of these distances all timesteps is referred to as the cumulative embedding variation (CEV). 

We find that the words with the lowest CEV are stopwords, e.g. \textit{from}, \textit{here}, \textit{there}. This makes sense, given that the embeddings are a function of how a word is used throughout a community, and these words are not expected to be impacted by the social evolution of the group. At the opposite end of the spectrum, showing the most variation, we observe less common words such as \textit{lackeys}, \textit{wooo}, \textit{fragment}, and \textit{booklet}. Noting that there seems to be a relation between word frequency and CEV, we calculate their Spearman correlation, yielding strongly negative value, $\rho=-0.77$ ($P<<0.05$).

Figure \ref{fig:word-change} shows the relation between the log frequency of a word and CEV, for all words in the training data (grey dots). We also plot the confirmed in-group language along the same axes (colourful dots) in Figure \ref{fig:word-change}. The strong negative correlation only appears to hold for frequencies under approx.\ 10~000; beyond this point, changes become close to zero. We have two hypotheses for why this might be. Firstly, it may simply be a computational issue, whereby words that are less common do not have high-quality representations due to insufficient training data, and therefore their embeddings are less stable. The second option is that words that are less common have more pliability in their colloquial usage. We expect the reality is a combination of the two.

We further observe that the variance in CEV covaries with frequency, such that a larger range of values is observed for low-frequency words. The confirmed in-group language tend to exhibit lower CEV per frequency, relative to the general vocabulary. Dotted lines indicate straight-line regressions to CEV of the whole vocabulary (black) and to the in-group language lexicon (yellow). We observe that the gradient for the in-group language is indeed less steep. This suggests that in-group language terms are more stable and less pliable in how they are used within this community, relative to what we would expect given their frequency. This could be a result of their special status as a social artefact in the group: there is an incentive to use the term ``correctly'' to signal group belonging.  

The different colours represent the predictions made by the best models in Section \ref{sec:results}, showing the top 1000 words predicted by the NPMI-category and \texttt{LISTN-C} (cluster-5) methods. The two methods appear to favour different areas in the frequency distribution, with the NPMI-based model being biased to the lower end of the spectrum, and the LISTN-based model to the mid and higher side. Future work may investigate combining these methods for lexicon induction.

Investigating the low-frequency, low-variation words, we find that it includes several mentions of medications (e.g.\ \textit{lamictal}, \textit{pristiq}, \textit{seroquel}). These are concepts with specific technical meanings that are not likely to vary due to socio-temporal changes in the community. We also note the term \textit{thotaudit}, which is the hashtag of a harassment campaign targeting sex workers. On the other end of the spectrum, \textit{submission} and \textit{bimbo} display a large amount of variation relative to their frequency. A possible reason for this might be that these terms occupy different niches for the subcommunities of the manosphere, or that they underwent a socio-semantic change during the training period. While more exploration of this embedding space is warranted, it is evident that it captures some interesting and useful qualities of the data.

\subsection{Sub-group linguistic specialisation}
The results in Section \ref{sec:results} show that lexicon induction benefits from accounting for social sub-group structure, suggesting that there might be sub-group linguistic specialisation. In this section, we further investigate this idea. The hypothesis is: if manosphere categories share a similar vernacular but express it to varying degrees, we would expect there to be a (nonlinear) correlation between their preference for different words in the lexicon. If the groups favour disjoint parts of the lexicon, there would be no such correlation.

\begin{table}[]
    \centering\footnotesize{
    \begin{tabular}{|p{0.17\linewidth}|p{0.75\linewidth}|}
    \hline
    \textbf{Group} & \textbf{Top words}\\
    \hline
    MGTOW &\textit{thot}, \textit{awalt}, \textbf{\textit{gyow}}, \textit{thots}, \textit{simps}, \textit{simp}, \textit{beta}, \textit{mgtow}, \textit{\textbf{cucked}}, \textit{alpha}, \textit{cuck}, \textit{bitches}\\
    \hline
    Incels& \textit{normie}, \textit{blackpill}, \textit{jfl}, \textit{volcel}, \textit{suifuel}, \textit{\textbf{braincels}}, \textit{normies}, \textit{foids}, \textit{\textbf{incels}}, \textit{foid}, \textit{\textbf{blackpilled}}, \textit{stacy}\\
    \hline
    Mental health& \textit{hurt}, \textit{failure}, \textit{personality}, \textit{killing}, \textit{cope}, \textit{harm}, \textit{kill}, \textit{hit}, \textit{tbh}, \textit{sa}, \textit{hurt}, \textit{dumb}, \textit{beat}\\
    \hline
    TRP &\textit{trp}, \textit{smv}, \textit{rp}, \textit{ltr}, \textit{oneitis}, \textit{alpha}, \textit{beta}, \textit{awalt}, \textit{redpill}, \textit{bp}, \textit{\textbf{asktrp}}, \textit{\textbf{hamstering}}\\
    \hline
    MRA &\textit{rape}, \textit{raped}, \textit{assault}, \textit{patriarchy}, \textit{female}, \textit{abused}, \textit{\textbf{mras}}, \textit{dumb}, \textit{hurt}, \textit{mgtow}, \textit{raping}, \textit{fgm}\\
    \hline
    PuA &\textit{ioi}, \textit{trp}, \textit{smv}, \textit{ltr}, \textit{oneitis}, \textit{alpha}, \textit{\textbf{iois}}, \textit{personality}, \textit{beta}, \textit{\textbf{seddit}}, \textit{\textbf{daygame}}, \textit{gf}\\
    \hline
    Criticism &\textit{trp}, \textit{\textbf{incels}}, \textit{manosphere}, \textit{alpha}, \textit{beta}, \textit{rp}, \textit{dumb}, \textit{personality}, \textit{asshole}, \textit{redpill}, \textit{redpilled}, \textit{mgtow}\\
    \hline
    \end{tabular}}
    \caption{Highest relevance lexicon terms per category.}
    \label{tab:group_terms}
\end{table}

\begin{figure}
    \centering
    \includegraphics[width=0.9\linewidth]{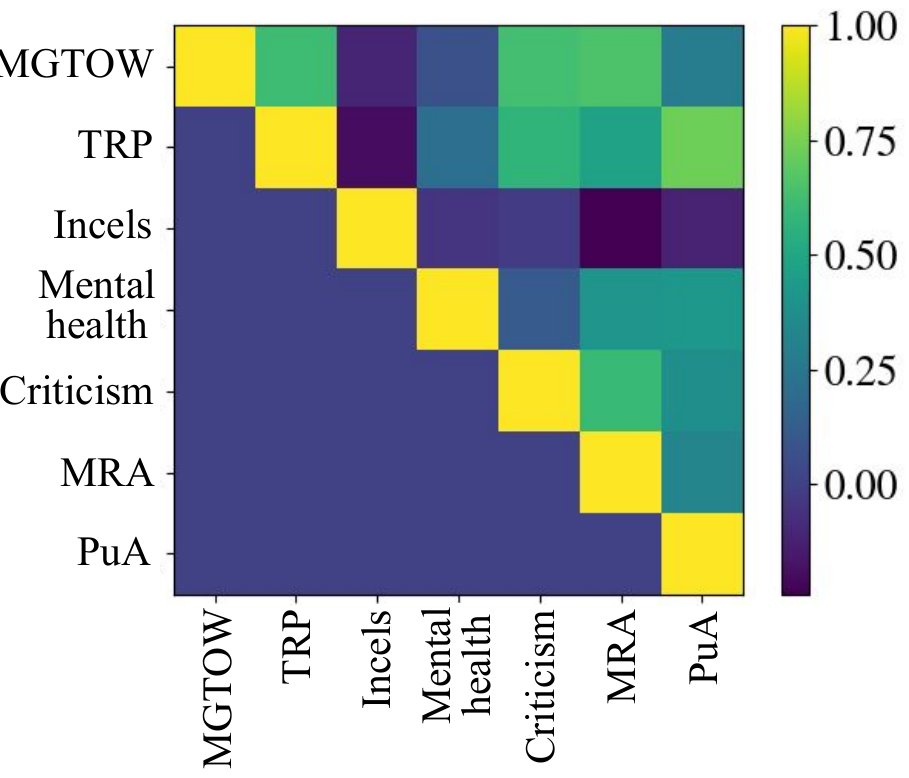}
    \caption{Spearman correlations between in-group language of different manosphere sub-groups.}
    \label{fig:cat-lang}
\end{figure}

To evaluate this idea, we firstly calculate the relevance of individual words in the lexicon to each of the manosphere categories. Table \ref{tab:group_terms} shows the highest-ranking lexicon words for each grouping. These words align with their known thematic interests; for example, the Pick-up Artists use terms such as \textit{IOI} (``indicators of interest'') and \textit{SMV} (``sexual market value''). To enable future research on these communities, these scores are also published.

We then find the Spearman correlation between the per-word relevance scores for each pair of communities, shown in Figure \ref{fig:cat-lang}. We observe that they correlate to varying degrees, with the largest correlation ($\rho=0.729$) between the Pick-up Artists (PuA) and The Red Pill (TRP). This has been observed in prior work in sociology: \citet{bachaud2024} states that PuA terminology is plentiful in the TRP community, as they have a shared focus on seduction. The MGTOWs have a fairly strong correlation with TRP ($\rho=0.615$) and MRA ($\rho=0.654$). This aligns with their historical connections: \citet{bachaud2024} states that MGTOW was created by a group of MRAs and that their ideology is ``close to both Red Pillers (critical about female nature) and MRAs (denouncing feminism and structural biases against men)''. A much weaker correlation is observed between MGTOW and PuA ($\rho=0.282$), which can be explained by their conflicting ideologies: PuAs are intent on pursuing women, whereas the main premise of MGTOW is to eschew contact with women. This captures an interesting insight into the relationship between PuAs, MGTOWs and TRPs: non-overlapping parts of the TRP interests are shared by both other groups.

The Incels have negative correlations with all other communities (with a minimum $\rho=-0.240$ compared to the Men's Rights Activists), showing that they have the most specialised in-group language. This, too, is supported by manosphere history: while the other manosphere groups have fairly intertwined origin stories, the Incels emerged from a separate line of new online platforms such as LoveShy (whose administrators openly supported mass killers and femicide) with cross-pollination from 4/chan and Reddit \citep{bachaud2024}. 

The Criticism groups, who are reformed manosphere members, correlate positively with MGTOW, MRA, TRP and PuA. Investigating their most relevant words (Table \ref{tab:group_terms}), we note that it mostly lists references to the manosphere itself.

\section{Conclusion}
In this paper, we introduced a novel framework for lexicon induction based on dynamic word and user embeddings. We compare several approaches based on two different basic models to methods from prior work, finding that the proposed methods achieve better scores on a newly crafted, expert-validated test set. A set of 455 new manosphere terms are to be released based on this analysis, with relevance scores for each manosphere sub-group. Our work provides novel insights into in-group language, including its embedding stability. The investigation into sub-group linguistic specialisation provides insights into the relationships between the manosphere subcommunities, which corroborate claims from prior work in sociology.

To our knowledge, this is the first work to incorporate social and temporal information to induce lexicons of in-group language. As the methodology is not specific to a particular group or language, we hope that it will be applied to create lexical resources for other communities in future. 

\section*{Ethical considerations}
The experiments in this paper make use of data developed in prior work. \citet{ribeiro2021evolution} state that the ethical standard guidelines of \citet{rivers2014ethical} were followed in the construction of this dataset, which includes not making any attempt to de-anonymize or link users across platforms. Even so, we recognise that the users did not consent to having their data analysed for the purpose of scientific experiments. However, given the real and current threat posed by this group, we view it as an acceptable use of the data. To limit the exposure of individuals, we do not provide any quotations or usernames, and all analyses are performed at an aggregate level.

As we are mindful of potential mental health repercussions of this work to collaborators, participants (including annotators) who were exposed to texts originating from manosphere communities were encouraged to attend a support group for researchers who work with extreme content.

\section*{Limitations}
There are a number of simplifying assumptions that are made in this work. All models evaluated (\texttt{LISTN} and baselines) provide only one score per word, and hence do not account for sense variation or polysemy. We further include only single-word lexical variations in our evaluation and exclude non-alphanumeric characters. Our motivation for these choices is provided in Section \ref{sec:evaluation}.

The \texttt{LISTN} and \texttt{word2vec} models are trained on data from 2018. Given the dynamic nature of in-group language, this limits the applicability of the lexicon to current data; however, the evaluations in this work serve to validate our lexicon induction approach. Given recent limitations placed on Reddit data access, the \citet{ribeiro2021evolution} dataset still constitutes one of the largest and most varied existing manosphere datasets. Future work may look at applying these methods to more modern datasets (possibly in the context of specialised forums like \url{incels.is}).

\section*{Acknowledgements}
This work is supported by the Hallmark Research Initiative on Fighting Harmful Online Communication. The author would like to thank Macken Murphy, Ed Hovy, and Indigo Orton for their valuable inputs. 

\bibliography{custom}

\appendix
\section{Hyperparameters}\label{app:hypers}
Hyperparameters for the representation learning component are shown in Table \ref{tab:hypers}.

\begin{table}[h!]
    \centering
    \begin{tabular}{|l|l|}
    \hline
    \textbf{Parameter} & \textbf{Value}\\
        \hline
        Early stopping patience & 5 epochs \\
        Early stopping tolerance & 0.001\\
        K&100\\
        Batch size & 80\\
        Weight of A&1\\
        Weight of C&1\\
        Learning rate&0.001\\
        $\lambda_1$&1\\
        $\lambda_2$&1\\
        $c_0$ scaler&0.01\\
        Max epochs&100\\
        \hline
    \end{tabular}
    \caption{Hyperparameters for training the Cerberus system.}
    \label{tab:hypers}
\end{table}

\section{Preliminary evaluation}\label{app:s1}
Results for the first step of the evaluation are shown in Table \ref{tab:res.s1}.

\begin{table}[h!]
    \centering
    \begin{tabular}{|l|l|l|}
        \hline & \textbf{AP} & \textbf{AUROC} \\
        \hline
        \multicolumn{3}{|c|}{\textbf{Baselines}}\\
        \hline
        PPMI &0.2162 & 0.3576\\
        PPMI-subreddit &0.2564 &0.4293 \\
        PPMI-category &0.2103 &0.3470 \\
        NPMI &0.2864&0.4171\\
        NPMI-subredit & \textbf{0.3377}&\textbf{0.5322} \\
        NPMI-category & 0.2748& 0.4114\\
        \hline
        \multicolumn{3}{|c|}{\textbf{LISTN-CA}}\\
        \hline
        Community centroid  &\textbf{0.6147}&\textbf{0.7344}\\
        Category centroid &0.5577&0.6997\\
        Subreddit centroid &0.3987&0.6287\\
        Cluster-5 &0.4302&0.637\\
        Cluster-100&0.3399&0.5554\\
        Bootstrap &0.2811&0.4960\\
        Bias &0.5690&0.7065\\
        \hline
        \multicolumn{3}{|c|}{\textbf{LISTN-C}}\\
        \hline
        Community centroid &0.6147 &0.7288\\
        Category centroid&\textbf{0.6809}&\textbf{0.7272}\\
        Subreddit centroid&0.5461&0.5891\\
        Cluster-5 &0.491&0.6827\\
        Cluster-100&0.4007&0.6184\\
        Bootstrap &0.5276 &0.5077\\
        Bias & 0.6015&0.5382\\
        \hline
        \end{tabular}
§        \caption{Results for preliminary experiments, used to construct the gold label test set.}
        \label{tab:res.s1}
    \end{table}
\end{document}